%% file: main.tex
\documentclass[10pt,twocolumn,letterpaper]{article}
\PassOptionsToPackage{table}{xcolor} 

\usepackage[pagenumbers]{cvpr} 

\usepackage{wrapfig}       
\usepackage{booktabs}      
\usepackage{graphicx}      
\usepackage{xcolor}       

\usepackage{graphicx}
\usepackage{booktabs}
\usepackage{amssymb}

\input{preamble}
\definecolor{cvprblue}{rgb}{0.21,0.49,0.74}
\usepackage[pagebackref,breaklinks,colorlinks,allcolors=cvprblue]{hyperref}

\def\confName{CVPR}
\def\confYear{2027}

\title{3D Tracking Any Point with Events and Frames}
\author{Roberto Pellerito\textsuperscript{1,*}\quad
Zijia Dai\textsuperscript{2,*}\quad
Laurent Kneip\textsuperscript{2}\quad
Davide Scaramuzza\textsuperscript{1}\\[0.5em]
{\textsuperscript{1}Robotics and Perception Group, University of Zurich}\\
{\textsuperscript{2}Mobile Perception Lab, ShanghaiTech University}
}

\begin{document}
\maketitle
\input{fig/teaser}
\begingroup
\renewcommand{\thefootnote}{}
\footnotetext{\textsuperscript{*}Equal contribution. This work was carried out under a programme of, and co-funded by, the European Space Agency (ESA), under Cooperative Agreement No. 4000150780/25/NL/GLC, with co-funding from Thales Alenia Space France. The views expressed herein can in no way be taken to reflect the official opinion of the European Space Agency..}
\endgroup
\input{sec/0_abstract}

\section*{Multimedia Material}
\textbf{Project Page:} https://rpg.ifi.uzh.ch/trackanymotion/

\input{sec/1_intro}
\input{sec/2_related_works}

\input{sec/3_method}
\input{sec/4_experiments}
\input{sec/5_conclusion}
\clearpage
{
    \small
    \bibliographystyle{ieeenat_fullname}
    \bibliography{main,supp}
}
\input{sec/X_suppl}

\end{document}

%% file: preamble.tex
\usepackage{cuted} 

\usepackage{currfile} 

\usepackage{caption} 

%% file: fig/teaser.tex
\begingroup
\preCutedStrip{\vspace{-1.8cm}}
\begin{strip}
    \centering
    \captionsetup{type=figure}
    \includegraphics[width=\textwidth]{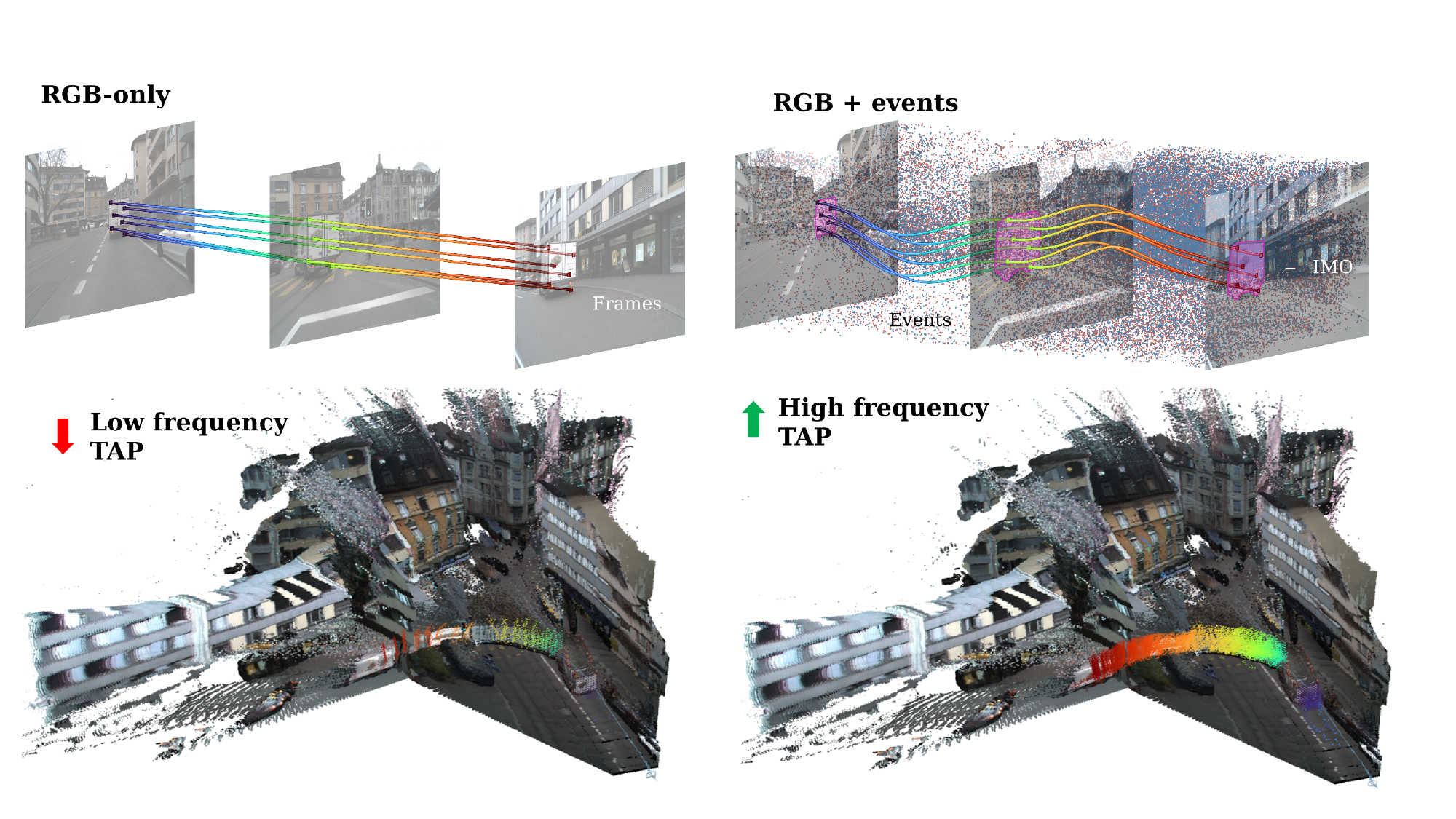}
    \caption{
          \textbf{Tracking any point in 3D, including between RGB frames.}
          TrackAnyMotion takes RGB frames and asynchronous events as input (right).
          It estimates dense 3D pointmaps, Indipendently moving objects (IMOs) and scene flow between pairs of timestamps (bottom), providing 3D correspondences across time.
          These correspondences are used to recover point trajectories at and between RGB timestamps at a greater time resolution compare to RGB-only methods (left), with color encoding time.
      }
    \label{fig:teaser}
\end{strip}
\endgroup

%% file: sec/0_abstract.tex

\begin{abstract}
Tracking a fast-moving point in 3D is prohibitive when RGB frames are blurred or too far apart to reveal its motion.
Event cameras record brightness changes between frames, providing measurements inside these temporal gaps.
We introduce TrackAnyMotion, which combines monocular RGB frames and events to track arbitrary points in 2D and 3D, including at times when no RGB frame is available.
The motion expected for a static point, given scene geometry and camera motion, provides a reference for identifying independently moving objects.
An independent-motion decoder combines geometric motion cues with event guidance and 2D tracks, feeding its predictions back into iterative 2D and 3D refinement.
We also introduce HyperEvent, a synthetic RGB--Event dataset with high-frequency 3D trajectories, geometry, and independent-motion labels.
At RGB frame timestamps on this benchmark, TrackAnyMotion reduces 3D and 2D endpoint errors by 70\% and 32\%, respectively. Our code and the newly introduced dataset HyperEvent will be released upon acceptance.
\end{abstract}

%% file: sec/1_intro.tex
\section{Introduction}
\label{sec:intro}
Tracking any point in 3D is fundamental for a vast variety of tasks spanning from robot imitation and policy learning~\cite{vecerik2024robotap, wen2023any}, dynamic reconstruction~\cite{wang2025shape, lei2025mosca}, motion-controlled video generation~\cite{geng2025motion, gu2025diffusion}, and physical inference~\cite{jiang2025phystwin, chen2026empm, wu2026event}. 

Despite the recent success of 3D point trackers, these methods are fundamentally limited by the hardware or standard cameras. A point on a fast-moving object travels far in the blind-time between consecutive frames, inevitably motion-blurring and corrupting its own observations. Moreover, standard cameras notably suffer from a bandwidth-latency trade-off that limits their performance in such cases. At low bandwidth and low frame rates, fast-objects appear motion blurred, while at high frame rates, motion blur is minimized at the expense of high bandwidth.


Event cameras have been shown to be an ideal complement to standard cameras in addressing the bandwidth-latency trade-off~\cite{gallego2022survey}.
They are bio-inspired sensors that asynchronously measure per-pixel brightness changes with microsecond temporal resolution. Due to their working principle, they trigger events sparsely, often in correspondence of moving edges yielding minimal bandwidth and providing information between subsequent frames.

Recent frame--event trackers such as TAPFormer~\cite{liu2026tapformer}
exploit event-based motion, demonstrating high-frequency 2D tracking between frames.
Yet until now, there exists no event-based 3D point tracker that leverages a set of monocular observations of events and frames to estimate the displacement of each point in a scene with high temporal resolution.
Moreover, existing methods typically resort to tracking a sparse set of points using complex hardware setups involving multiple cameras



To overcome existing methods, we propose the first 3D track-any-point method, which leverages the high temporal resolution of event cameras in combination with standard frames to estimate 2D and 3D point displacement at high frequency. 

We use the expected motion for a static point as the reference for tracking and disentangling motion caused by independently moving objects and by the static background.
By fusing frames and event features and iteratively conditioning 2D and 3D updaters with event features, we are able to extrapolate the geometry of the scene as well as the 2D-3D point displacement with high frequency.



With this work, we introduce \textbf{HyperEvent} , a large scale simulated dataset consisting of high frequency ground truth of depth, 2D-3D tracks, poses, and interdependently moving objects. with events and frames.
To the best of our knowledge, this is the first dataset tailored for 3D tracking and independently moving object segmentation using events and frames.

Our tracker outperforms state-of-the-art baselines in both 2D and 3D tracking by 70\% and 32\%, respectively. Moreover, TrackAnyMotion enables scenarios that are inaccessible to standard frame-based cameras by tracking points at arbitrary timestamps in time, outperforming interpolation-based baselines by 2x in both 2D and 3D.

With this work we make the following contributions:
\begin{enumerate}
    \item We present \textbf{TrackAnyMotion}, a monocular RGB--event tracker that follows arbitrary query points in 2D and 3D, including at timestamps between RGB frames (\cref{sec:event,sec:seed})
    \item We introduce geometry-conditioned independent-motion reasoning within iterative tracking, using deviations from camera-induced motion to predict motion masks and feed their features back into 2D and 3D refinement (\cref{sec:seed,sec:imo})
    \item We construct \textbf{HyperEvent}, a synthetic RGB--Event dataset with high-frequency 3D trajectories, geometry, and independent-motion labels for training and controlled evaluation.
\end{enumerate}

%% file: sec/2_related_works.tex
\section{Related work}
\label{sec:related_work}

\paragraph{Point tracking and dynamic 3D reconstruction.}
Image-space trackers such as CoTracker3~\cite{karaev2025cotracker3} learn correspondences from synthetic supervision and pseudo-labelled real videos, and a growing family of methods lifts these correspondences into 3D. SpatialTracker~\cite{xiao2024spatialtracker}, DELTA~\cite{ngo2025delta}, TAPIP3D~\cite{zhang2025tapip3d} and TrackingWorld~\cite{lu2025trackingworld} rely on depth estimated by a separate network, whereas SpatialTrackerV2~\cite{xiao2025spatialtrackerv2}, St4RTrack~\cite{feng2025st4rtrack}, D4RT~\cite{zhang2026d4rt} and V-DPM~\cite{sucar2026vdpm} predict scene geometry and point motion jointly in a single feed-forward pass. Trace Anything~\cite{liu2026traceanything} represents every trajectory as a B-spline in time, so that positions can be read off at any instant, and TrackCraft3R~\cite{nam2026trackcraft3r} and ReLift~\cite{singer2026relift} start from the output of a reconstruction model and refine tracks on top of it. TrackEverything~\cite{jain2026trackeverything} extends dense tracking to videos of more than a thousand frames with a persistent world-space representation. Our tracking backbone is Track4World~\cite{lu2026track4world}, which predicts world-centric dense 2D and 3D flow and refines it with iterative updates in the style of RAFT~\cite{teed2020raft}. All of these methods see only RGB frames and report positions at frame times. TrackAnyMotion adds the event stream of the same camera and predicts 2D and 3D positions in between frames.

\paragraph{Tracking with events and frames.}
Event cameras report per-pixel brightness changes asynchronously and with microsecond latency, so they observe how the scene moves while a frame camera is blind between exposures~\cite{gallego2022survey}. Early data-driven trackers followed frame-initialised features with events alone~\cite{messikommer2023featuretracking}, and ETAP~\cite{hamann2025etap} and MATE~\cite{han2025mate} extended this to tracking any point from an event stream. Combining both modalities, FE-TAP~\cite{liu2024fetap}, TAPFormer~\cite{liu2026tapformer} and BlinkTrack~\cite{shen2025blinktrack} fuse frames and events to output 2D tracks well above the frame rate, and our event encoder follows the design of TAPFormer. Continuous-time formulations go one step further and model each pixel's trajectory as a parametric curve that can be sampled at any time~\cite{gehrig2024densecontinuous,wan2025edcpt}, but the trajectories remain in the image plane. In 3D, EMoTive~\cite{wan2025emotive} and Han et al.~\cite{han2026unsup3dmotion} recover optical flow together with motion in depth from events, which yields a 3D motion field over a short horizon rather than persistent correspondences. The event-based trackers that do produce 3D trajectories need additional sensing: stereo events in E-3DTrack~\cite{li2024event3d}, a rig of four RGB and two event cameras in MER-Tracker~\cite{chang2026mertracker}, or a depth map as input in E-TraMamba~\cite{shen2026etramamba}. We work with a single RGB camera and its co-located events, take no depth as input, and output long-range 3D tracks of arbitrary points at timestamps between frames.

\paragraph{Visual geometry models with event information.}
Feed-forward geometry models such as VGGT~\cite{wang2025vggt}, $\pi^3$~\cite{wang2026pi3} and Depth Anything 3~\cite{lin2025da3} predict depth, pointmaps and camera parameters directly from an arbitrary set of views, MonST3R~\cite{zhang2025monst3r} adapts this recipe to scenes with moving objects, and Flow3r~\cite{cong2026flow3r} shows that correspondence supervision can drive geometry and pose learning. Events have recently entered this family from two directions. Depth AnyEvent~\cite{bartolomei2025depthanyevent} and EventVGGT~\cite{ren2026eventvggt} distill RGB depth and geometry models into event-based estimators, while EAG3R~\cite{wu2025eag3r} and Weerakoon et al.~\cite{weerakoon2026eventsrgb4d} add an event branch to pointmap reconstruction for dynamic scenes and difficult lighting, and Interp3R~\cite{guo2026interp3r} uses events to interpolate pointmaps and camera poses to timestamps where no frame exists. These models produce geometry at each timestamp but do not link material points across time. We rely on them for exactly that geometry: Depth Anything 3 provides the depth and camera poses that serve as the static-scene reference for our tracking and for the rigid-flow computation described below.

\paragraph{Independently moving objects segmentation.}
Separating the motion a camera induces from the motion of objects is a long-standing problem, and the most common tool is a comparison between the observed motion and the rigid flow that camera motion alone would produce given the scene depth. GeoNet~\cite{yin2018geonet} introduced this decomposition for unsupervised depth and pose learning, treating the residual as object motion. Event-based methods adopted the same idea in several forms: through motion compensation and contrast maximization~\cite{stoffregen2019motionseg}, by learning depth, ego-motion and motion masks jointly~\cite{mitrokhin2019evimo}, or, in Un-EVIMO~\cite{wang2024unevimo}, by thresholding the rigid-flow residual into pseudo-labels for an event segmenter. More recent learned segmenters feed ego-motion-compensated events and temporal attention to the network~\cite{georgoulis2024outofroom}, predict masks together with future motion to suppress events on moving objects~\cite{pellerito2026suppression}, or supervise boundary ownership alongside motion masks~\cite{hua2026borderownership}. On the RGB side, DynOSeg~\cite{han2027dynoseg} fuses a geometry-aware residual map with depth features, and Segment Any Motion~\cite{huang2025segmentanymotion} reads moving objects off long-range tracks. Trackers themselves have started to carry motion structure: SpatialTracker~\cite{xiao2024spatialtracker} and MoSE3~\cite{cheng2026mose3} learn rigidity embeddings for trajectories, and TrackEverything~\cite{jain2026trackeverything} classifies points as static or dynamic and refines only the dynamic ones. What these approaches share is that the residual, or the static-versus-dynamic decision, is computed once, as preprocessing, as pseudo-labels, or as a single routing step. In TrackAnyMotion the residual is a running quantity instead: we recompute it from the current depth, pose and track estimates inside every refinement iteration, and use it both to segment independently moving objects (IMOs) and to decide how much each update should trust the event evidence.

%% file: sec/3_method.tex
\begin{figure*}[t]
    \centering
    \includegraphics[width=\linewidth]{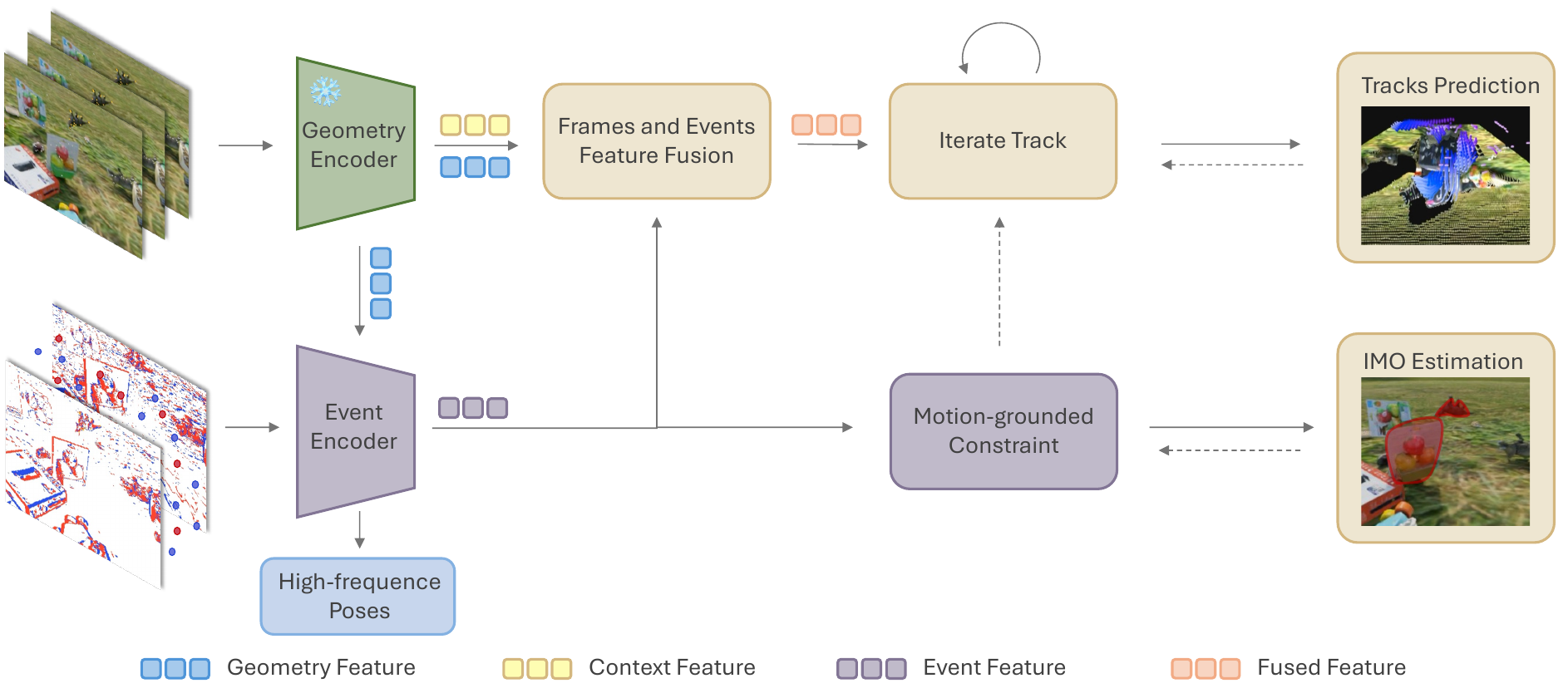}
    \caption{\textbf{Overview of TrackAnyMotion.}
    Frozen RGB geometry and event-conditioned features support iterative 2D and 3D tracking.
    At inter-frame timestamps, learned camera poses and coarse world motion construct target pseudo-pointmaps.
    An independent-motion decoder combines tracking features with deviations from camera-induced motion; its features and probabilities feed back into both trajectory updates.}
    \label{fig:pipeline}
\end{figure*}
\section{Method}
\label{sec:method}

Our method jointly estimates 2D and 3D point trajectories and independent-motion masks from monocular RGB frames and synchronized events.
Our central idea is to use camera-induced motion as a geometric reference within iterative tracking: deviations from this reference help identify independently moving points, whose learned motion features in turn condition trajectory refinement.
For timestamps without an RGB observation, event-conditioned pose and motion predictions supply the target geometry needed by the tracker.
\cref{fig:pipeline} summarizes these two mechanisms.

\subsection{Problem Setup}
\label{sec:setup}

\paragraph{Input and output.}
The inputs are RGB frames $\{I_s,t_s\}_{s=1}^{S}$ and events $\mathcal{E}=\{(\mathbf{x}_j,t_j,p_j)\}$, with pixel location $\mathbf{x}_j$, timestamp $t_j$, and polarity $p_j\in\{-1,+1\}$.
We predict at RGB timestamps and requested inter-frame times $t_k$.
For bracketing RGB times $t_k^-,t_k^+$, define $\alpha_k=(t_k-t_k^-)/(t_k^+-t_k^-)$.
Given source point $\mathbf{u}_{s,i}$, the tracker predicts target image position $\widehat{\mathbf{u}}_{t,i}$, visibility, confidence, and a camera-coordinate residual:
\begin{equation}
  \widehat{\mathbf{P}}_{t,i}
  =\widehat{\mathbf{P}}_{s,i}+\widehat{\mathbf{r}}_{s\to t,i}.
  \label{eq:target_pos}
\end{equation}
The endpoints use their respective camera frames, so this residual includes camera motion.
We additionally predict independent-motion probability $p_{t,i}$ on the source-point grid.
Superscript $n$ denotes refinement iteration and $\mathrm{gt}$ denotes ground truth.

\paragraph{Base tracker.}
We build on Track4World~\cite{lu2026track4world}, using frozen DA3~\cite{lin2025da3} geometry to obtain RGB pointmaps and camera-to-world poses $\widehat{T}_s=(\widehat{R}_s,\widehat{\mathbf{t}}_s)$.
Its recurrent decoder refines 2D and 3D residuals using appearance, context, and pointmap correlations.
The geometry backbone stays frozen while tracking components and the added event and IMO modules are trained.

\subsection{Event-Conditioned Query Features}
\label{sec:event}

\paragraph{Encoding and temporal fusion.}
At each query timestamp, we form a ten-channel stack~\cite{hamann2025etap} event representation from up to 400,000 preceding events, with successively shorter count-based windows.
A TAPFormer event encoder~\cite{liu2026tapformer} maps the stack to stride-8 features $\mathbf{E}_k$.
Temporal fusion maintains a transient state: at RGB observations it is reset from RGB context, then updated with the query's events through a convolutional GRU.
A local $7\times7$ cross-attention uses source RGB context to query this state, producing an event correction $\boldsymbol{\delta}_k$ through a learnable channel gate.

\paragraph{Query appearance.}
For an inter-frame query $t_k$, stream $a$ (appearance, context, or 3D descriptors) combines bracketing RGB features with an event correction:
\begin{equation}
 \widetilde{\mathbf{F}}_k^a
 = (1-\alpha_k)\mathbf{F}_{k^-}^a
 +\alpha_k\mathbf{F}_{k^+}^a
 +\mathbf{g}_a\odot\psi_a(\boldsymbol{\delta}_k).
\end{equation}
Here $\mathbf{F}_{k^-}^a$ and $\mathbf{F}_{k^+}^a$ are the features at the preceding and following RGB frames, and $\alpha_k$ is the query's relative time between them.
The learned $1\times1$ convolution $\psi_a$ maps the event correction $\boldsymbol{\delta}_k$ to the stream's $C_a$ channels.
The learned vector $\mathbf{g}_a\in\mathbb{R}^{C_a}$ scales each correction channel; $\odot$ denotes channel-wise multiplication broadcast over spatial locations.
These gates are shared across queries and spatial positions, with unconstrained values learned during training.
RGB queries also receive event corrections at their corresponding slots.
The correction and injection gates start at zero, preserving the RGB features initially.
A fused context $\mathbf{F}_k$, formed by adding $\boldsymbol{\delta}_k$ to source RGB context, additionally conditions the temporal geometry heads.
Although event windows end at the query, the evaluated appearance and pose paths use both bracketing RGB frames and are therefore bidirectional.

\subsection{Geometry Between RGB Frames}
\label{sec:seed}

A 3D correspondence decoder needs target geometry even when no RGB frame exists at the requested time.
We construct a pseudo-pointmap by predicting camera pose and spatially varying world motion.

\paragraph{Event-conditioned pose.}
Let $\overline{T}_k$ interpolate the detached bracketing RGB poses, using SLERP for rotation and linear interpolation for translation.
A pooled fused feature and temporal phase predict a pose correction:
\begin{equation}
 \widehat{T}_k=\overline{T}_k\exp(\boldsymbol{\xi}_k^\wedge),
 \qquad
 \boldsymbol{\xi}_k=g_\theta(\mathrm{pool}(\mathbf{F}_k),\alpha_k).
\end{equation}
The learned twist adjusts the interpolated camera trajectory without updating the frozen RGB geometry.

\paragraph{Motion-conditioned pseudo-pointmaps.}
A convolutional head takes source RGB context and $\mathbf{F}_k$ to predict a coarse world-displacement field $\Delta\mathbf{X}_{k,i}^{w}$ on the stride-8 grid.
Each source point is moved in world coordinates and transformed into the query camera:
\begin{equation}
 \widetilde{\mathbf{P}}_{k,i}
 =\widehat{T}_k^{-1}
   \bigl(\widehat{T}_s\widehat{\mathbf{P}}_{s,i}
         +\Delta\mathbf{X}_{k,i}^{w}\bigr).
\end{equation}
Transforms act on 3D positions with translation included.
We project and forward-splat these points into the target grid, using soft depth weights to favor nearer surfaces where projections overlap.
Unfilled cells use the camera-only transformed source pointmap.
This pseudo-pointmap supplies the target geometric correlations, while $\widetilde{\mathbf{F}}_k^a$ supplies target appearance.
The coarse field is computed once per forward pass and detached from the warp; a separate displacement loss trains its head.
Thus camera motion and object motion both shape the geometry presented to iterative refinement.

\subsection{Independent-Motion Feedback}
\label{sec:imo}

\paragraph{Geometric reference.}
A static source point would appear at
\begin{equation}
 \mathbf{P}_{t,i}^{\mathrm{rig}}
 =\widehat{T}_t^{-1}\widehat{T}_s\widehat{\mathbf{P}}_{s,i},
 \qquad
 \mathbf{u}_{t,i}^{\mathrm{rig}}=\pi_t(\mathbf{P}_{t,i}^{\mathrm{rig}}),
 \label{eq:rigid_flow}
\end{equation}
where projection uses calibration estimated from predicted pointmaps.
After each 2D update, we measure deviation from this rigid projection and the world displacement implied by the previous 3D estimate:
\begin{align}
 \mathbf{a}_{t,i}^{(n)}
 &=\widehat{\mathbf{u}}_{t,i}^{(n)}-\mathbf{u}_{t,i}^{\mathrm{rig}},\\
 \mathbf{b}_{t,i}^{(n)}
 &=\widehat{T}_t\widehat{\mathbf{P}}_{t,i}^{(n-1)}
   -\widehat{T}_s\widehat{\mathbf{P}}_{s,i}.
 \label{eq:world_motion}
\end{align}
At the first iteration, the rigid target initializes this cue, giving zero world displacement.
Image residuals are normalized by grid size and world displacement by source-point distance; bounded motion cues are combined with source/target confidence and geometric validity.
These quantities provide evidence of independent motion despite camera movement and uncertain geometry.

\paragraph{Track-aligned IMO decoding.}
At the current correspondence, we sample target event features and the target 3D descriptor.
A shared spatial decoder combines them with source context, the current tracking state, the target-minus-source descriptor, and the geometric cues.
Bidirectional temporal attention then aggregates features along each source-anchored track, without mixing spatial locations in the attention operation.
The resulting feature $\mathbf{z}_{t,i}^{(n)}$ predicts an IMO logit $\ell_{t,i}^{(n)}$ and probability $p_{t,i}^{(n)}=\sigma(\ell_{t,i}^{(n)})$.
The final mask thresholds this probability at $0.5$.
Pairwise inference has one target slot, so temporal aggregation reduces to that slot.

\paragraph{Feedback into tracking.}
Two adapters convert the IMO feature and probability into residual tracking features:
\begin{equation}
 \Delta\mathbf{h}_{d}^{(n)}
 = A_d\bigl([\mathbf{z}^{(n)},p^{(n)}]\bigr),
 \qquad d\in\{2\mathrm{D},3\mathrm{D}\}.
\end{equation}
The 3D adapter conditions the current 3D update; the 2D adapter conditions the next recurrent 2D update.
Both adapters have zero-initialized output layers, initially preserving the pretrained updates.
Their feedback connections retain gradients across iterations.
Consequently, segmentation uses the evolving correspondence estimates and returns learned motion information to trajectory refinement, rather than acting only as a final output.

\subsection{Training Objective}
\label{sec:loss}

We train on RGB and inter-frame query batches, supervising tracking and IMO at all $N=4$ refinement iterations.
Geometry predictions remain frozen; all 3D supervision is expressed in the corresponding normalized geometry coordinates.

\paragraph{Tracking supervision.}
Let $\langle f\rangle_w=\sum_iw_if_i/\max(\sum_iw_i,\epsilon)$, with validity-masked visibility weights $w_i=v_i^{\mathrm{gt}}+0.2(1-v_i^{\mathrm{gt}})$.
The trajectory losses are
\begin{align}
 \mathcal{L}_{2\mathrm{D}}^{(n)}
 &=\left\langle\|\widehat{\mathbf{u}}_{t,i}^{(n)}
                  -\mathbf{u}_{t,i}^{\mathrm{gt}}\|_1\right\rangle_w,\\
 \mathcal{L}_{3\mathrm{D}}^{(n)}
 &=\left\langle\|\widehat{\mathbf{r}}_{s\to t,i}^{(n)}
                  -\overline{\mathbf{r}}_{s\to t,i}^{\mathrm{gt}}\|_1\right\rangle_w.
\end{align}
The target $\overline{\mathbf{r}}^{\mathrm{gt}}$ uses the training geometry-scale alignment.
Following the base tracker, visibility and confidence receive binary cross-entropy losses; confidence targets indicate 2D error below 3 pixels.
A squared spatial smoothness penalty regularizes physical displacement in source-camera axes over valid, same-instance geometric neighbors.
Denote their weighted sum by $\mathcal{L}_{\mathrm{track}}^{(n)}$, using unit tracking weights and smoothness weight $0.05$.

\paragraph{IMO supervision.}
Native source-material IMO annotations supervise logits sampled at the same source points across target times.
Unknown annotations, invalid correspondences, source-time queries, and padded slots are excluded.
A one-pixel boundary band softens targets using distances on the annotation grid; the remaining labels stay binary.
The loss is the mean binary cross-entropy $\mathcal{L}_{\mathrm{IMO}}^{(n)}$ over eligible points, without deriving labels from predicted motion.

\paragraph{Temporal geometry supervision.}
On inter-frame batches, $\mathcal{L}_{\mathrm{pose}}$ combines Frobenius rotation error and translation $L_1$ error, and $\mathcal{L}_{\mathrm{coarse}}$ supervises the coarse world-displacement field with $L_1$ error.
Ground truth is transformed into the predicted reference frame and normalized scale before these comparisons.
The complete objective is
\begin{equation}
\begin{split}
 \mathcal{L}={}&\sum_{n=1}^{N}\gamma^{N-n}
   \bigl(\mathcal{L}_{\mathrm{track}}^{(n)}
        +\mathcal{L}_{\mathrm{IMO}}^{(n)}\bigr)\\
 &+\mathbb{I}_{\mathrm{inter}}\bigl(\mathcal{L}_{\mathrm{pose}}
                  +0.5\mathcal{L}_{\mathrm{coarse}}\bigr),
\end{split}
\end{equation}
with $\gamma=0.8$ and $\mathbb{I}_{\mathrm{inter}}$ indicating the temporal query branch.

%% file: sec/4_experiments.tex
\section{Experiments}
We evaluate TrackAnyMotion, trained solely on HyperEvent, on high-frequency
trajectory interpolation, 2D and 3D point tracking, and independent-motion
segmentation. Tests on synthetic and real-world datasets and ablations
examine event input, motion feedback, and tracking at and between RGB timestamps.

\input{fig/qualitative_figure}

\paragraph{Implementation details.}
Our model builds on pretrained Track4World tracking components and DA3
geometry. At RGB timestamps, the pairwise inference protocol processes
each source--target pair independently using two RGB frames and four
refinement iterations in half precision. 

\paragraph{Datasets.}
HyperEvent contains 1,499 scenes, split into 1,199 training,
150 validation, and 150 test scenes. Across all 25,621 foreground objects,
the average relative speed is 1.458~m/s and the peak relative speed is
63.650~m/s. For the 3,029 dynamic objects, the average absolute
speed is 6.324~m/s and the peak absolute speed is 29.731~m/s.
The native evaluation uses
$256\times256$ RGB images and a $32\times32$ material-point grid, with RGB
observations at 50~Hz and geometric trajectories at 200~Hz. For external
tracking evaluation, we use 200 KITTI~2015 image pairs and the released
PointOdyssey and ADT mini subsets. PointOdyssey images are resized to
$640\times360$, while ADT uses its native $512\times512$ resolution.
PointOdyssey and ADT are evaluated using independent frame-0-to-target
pairs within 16- and 50-frame prefixes. Events are simulated for Kitti, ADT and PointOdissey datasets, while DSEC reports real events.

\subsection{High-frequency trajectory interpolation}
\label{sec:experiments_interpolation}

\input{tables/highfrequency}
We predict 200~Hz trajectories from 25~Hz RGB frames and intervening
events, including seven intermediate positions per 40~ms interval.
Table~\ref{tab:tapformer_comparison} compares RGB timestamps (S) and
intermediate timestamps (K) against linear interpolation (LI) of
Track4World and TrackEverything tracks, and TAPFormer~\cite{liu2026tapformer}
for 2D tracking.

At intermediate timestamps, TrackAnyMotion achieves 0.132~px EPE2D,
compared with 1.470~px for Track4World + LI and 0.497~px for TAPFormer.
Its EPE3D is 0.148~m, versus 0.284~m and 0.382~m for Track4World + LI
and TrackEverything + LI, respectively.
Gains also extend to RGB timestamps and both IMO and static regions.
These cross-model comparisons do not isolate the contributions of events
or motion grounding; matched controls are discussed in the ablation study.
The supplementary section provides the protocol and detailed results.

\subsection{2D point tracking}
\label{sec:experiments_2d}
\input{tables/tracking2d}

We measure 2D tracking accuracy using endpoint error in input-image pixels
and the percentages of predictions with errors strictly below 1 and
3~pixels. Point coordinates are converted to input-image pixels
before scoring. For our synthetic frame-time evaluation, eligible
correspondences have valid, visible, in-frame source and target
observations, finite ground-truth coordinates, and positive depth. Errors and accuracy counts are pooled over eligible correspondence occurrences.
Table~\ref{tab:tracking2d} compares EKLT, DeepEvT, TAPFormer, and
our method on fixed pilot cohorts of 494 HyperEvent and 256 DSEC
correspondence occurrences. Our method achieves the lowest EPE2D
and highest strict and relaxed accuracies on both datasets.
Relative to TAPFormer, EPE2D decreases from 1.224 to 0.728~px on
HyperEvent and from 1.932 to 1.605~px on DSEC.

\subsection{3D point tracking}

\input{tables/full-3D}

\input{tables/imo}
\input{tables/ablations}

We jointly align each pair of predicted source and target point clouds
to ground truth using one shared scale and translation per method and
pair. We report aligned 3D residual error ($\mathrm{EPE}_{3D}$), strict
and relaxed accuracies (AccS/AccR), and XYZ geometry metrics
(AbsRel/$\delta_{1.25}$). These scores evaluate camera-coordinate
residuals after alignment and do not establish absolute metric-scale
recovery. KITTI errors and absolute thresholds use the released,
uncalibrated coordinate scale. Exact computations and regional
aggregation are given in the supplementary material.

\subsection{Independently Moving Object Estimation}
We evaluate segmentation on 150 synthetic scenes using native
$256\times256$ ground-truth masks and a prediction threshold of 0.5.
We report mIoU and pIoU and averaging procedure in the supplementary material.
Event Suppression uses causal 50~ms event windows, while our method uses the source and next RGB frames with their event stacks.


\subsection{Ablation Study}
\label{sec:experiments_ablations}

We investigate tracking under sparse RGB observations using three variants:
\emph{T4W--TAPFormer} (A), which combines Track4World with TAPFormer
event encoding and fusion without IMO;
\emph{Ours} (B), which additionally introduces IMO prediction
and conditioning; and
\emph{T4W--ETAP} (C), which uses ETAP encoding and its associated
fusion without IMO.
We also evaluate \emph{Ours--ZeroEvents} (D), where we zeros event representation starting from B model while retaining its weights,
fusion connections, IMO branch, timestamps, and temporal query axis.

\paragraph{Setup and metrics.}
We evaluate three held-out scenes at 5\,Hz RGB and 200\,Hz trajectory
sampling, using two RGB frames 200\,ms apart and 41 query timestamps.
We report EPE2D (px) and aligned camera-residual EPE3D (m).
Details about sampling, alignment, eligibility, motion grouping, and timing
are available in the supplementary material.

\paragraph{Different Motion Levels.}
We compare static points with the medium- and high-motion GT-speed groups.
Table~\ref{tab:ablations}(a) reports RGB-endpoint tracking errors.
The largest event-input effect occurs in the high-motion group:
zeroing events increases EPE2D from 7.629 to 15.435\,px,
a 102.31\% increase.
The effect is not uniform across groups: the zero-event control
has lower static and medium-motion EPE2D.
These results indicate that event input primarily reduces large
2D tracking errors associated with high speed motion.

\paragraph{Accuracy Changes Between RGB Timestamps.}
Table~\ref{tab:ablations}(b) compares direct predictions with linear
interpolation (LI) of each model's own RGB endpoints at 25\%, 50\%, and
75\% of the inter-frame interval. Our endpoints are event-conditioned;
endpoint and intermediate evaluations use different GT-valid cohorts.

At the three reported phases, our method has lower native EPE2D
than its zero-event control, but higher aligned residual EPE3D.
Interpolation of our own endpoints outperforms our direct
predictions at all three reported phases for both errors.
Across all 39 phases, its interpolated predictions have lower
EPE2D in 37 phases and lower EPE3D in 38 phases.
Thus, the sparse-RGB results support the advantage of using event for 2D
tracking. 

C has no native learned intermediate output in the evaluated
integration and is reported as N/A.

\paragraph{Effectiveness of Components.}
Table~\ref{tab:ablations}(c) summarizes results on the common
RGB-endpoint cohort of 423 correspondences.

\emph{IMO modeling: B versus A.}
B adds an IMO decoder with feedback into 2D and 3D refinement,
using the architecture and supervision described in
Secs.~\ref{sec:imo} and~\ref{sec:loss}.
B has lower overall tracking errors than A in this pilot.

\emph{Event encoding and fusion: C versus A.}
C uses the ETAP encoder and gated residual adapters for dense 2D,
3D, and context features, replacing both A's event encoder and
its fusion mechanism.

\emph{Event input: D versus B.}
Zeroing events increases overall EPE2D from 1.961 to 3.136\,px
(+59.93\%) and IMO-region EPE2D from 3.472 to 6.040\,px.
However, overall aligned residual EPE3D slightly improves
(0.150 to 0.149\,m), as does the fraction of RGB-endpoint predictions
below one pixel (0.6028 to 0.7021).
Events therefore primarily reduce large 2D errors under high motion,
rather than uniformly improving tracking.

%% file: fig/qualitative_figure.tex
\begin{figure}[t]
    \centering
    \includegraphics[width=\linewidth]{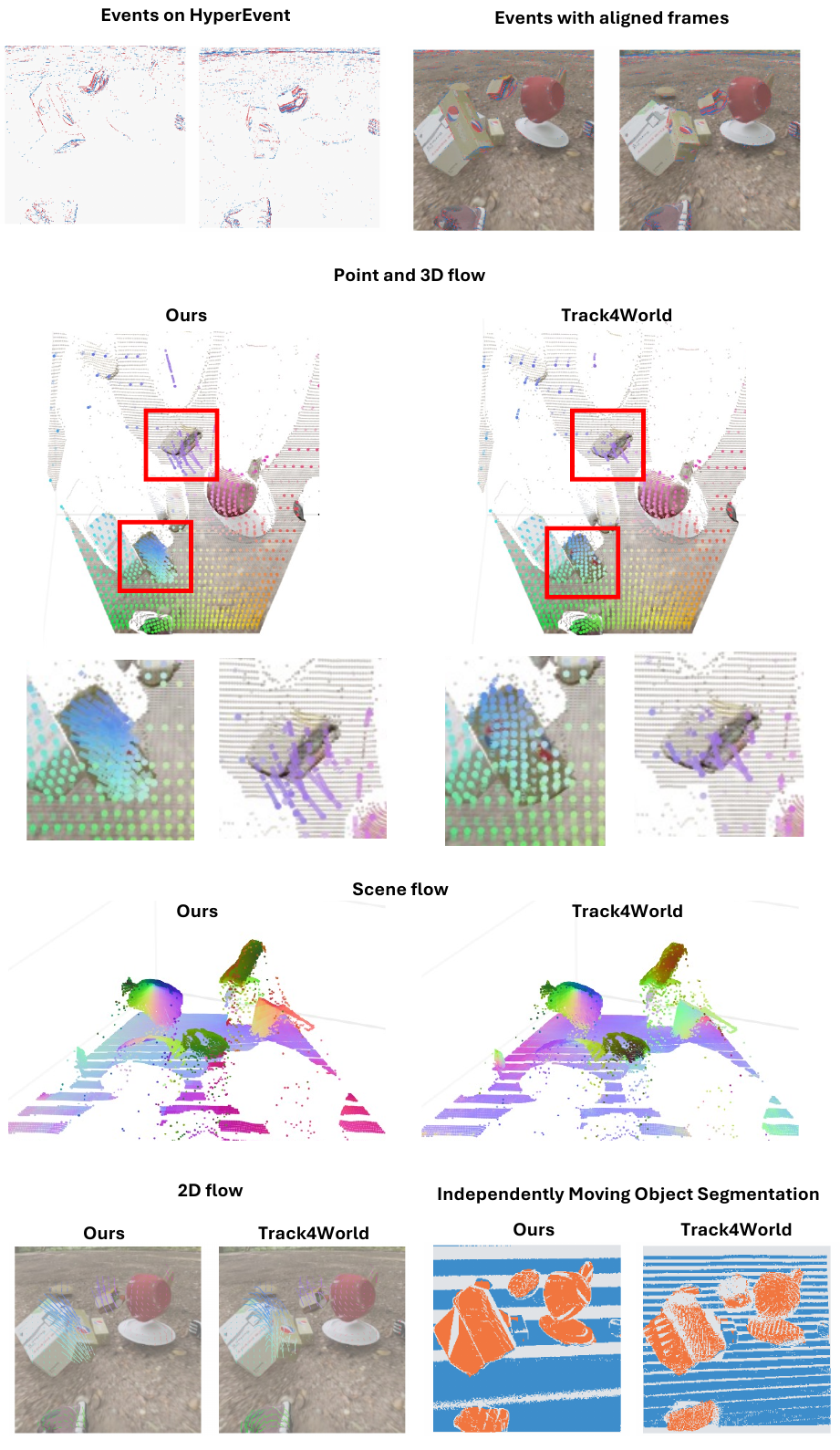}
    \caption{\textbf{Qualitative comparison on HyperEvent.}
    Top: events and their overlays on aligned RGB frames.
    TrackAnyMotion and Track4World are compared through 3D points and flow,
    scene flow, 2D flow, and independent-motion segmentation.
    Red boxes mark regions enlarged below.
    In the segmentation maps, orange denotes independently moving objects
    and blue denotes static regions.}
    \label{fig:qualitative}
\end{figure}

%% file: tables/highfrequency.tex
\begin{table*}[t]
    \centering
    \scriptsize
    \renewcommand{\arraystretch}{1.15}
    \caption{
        2D and 3D tracking at RGB observation timestamps (S axis) and
        bracketed intermediate timestamps (K axis).
        The 2D cohorts contain 560,615 and 153,519 correspondence occurrences,
        respectively. EPEs are in pixels for 2D and metres for 3D;
        accuracies are percentages. In 2D, AccS/AccR use strict 1/3-pixel
        thresholds; in 3D, they use absolute-or-relative thresholds of
        0.05/0.10. Regional EPEs use valid source-frame IMO/static labels;
        static includes stationary objects and background.
        LI denotes linear interpolation between predicted RGB endpoints,
        used by both Track4World and TrackEverything on the K axis.
        Bold indicates the best values within each block.
    }
    \label{tab:tapformer_comparison}
    \setlength{\tabcolsep}{3pt}
    \resizebox{\linewidth}{!}{%
    \begin{tabular}{@{}l ccccc ccccc@{}}
        \toprule
        \textbf{Method}
        & \multicolumn{5}{c}{\textbf{S axis}}
        & \multicolumn{5}{c}{\textbf{K axis}} \\
        \cmidrule(lr){2-6} \cmidrule(lr){7-11}
        & EPE $\downarrow$ & AccS $\uparrow$ & AccR $\uparrow$
        & IMO EPE $\downarrow$ & Static EPE $\downarrow$
        & EPE $\downarrow$ & AccS $\uparrow$ & AccR $\uparrow$
        & IMO EPE $\downarrow$ & Static EPE $\downarrow$ \\
        \midrule
        \multicolumn{11}{@{}l}{\textbf{2D tracking (EPE in pixels)}} \\
        \midrule
        Track4World~\cite{lu2026track4world} + LI
        & 1.219 & 68.38 & 91.50 & 1.950 & \textbf{0.637}
        & 1.470 & 62.38 & 89.32 & 2.492 & 0.913 \\
        \rowcolor{gray!20}
        TrackEverything~\cite{jain2026trackeverything} + LI
        & 2.695 & 40.52 & 83.91 & 4.252 & 2.139
        & 2.466 & 45.82 & 85.60 & 4.047 & 1.910 \\
        TAPFormer~\cite{liu2026tapformer}
        & 1.230 & 84.25 & 94.90 & 2.063 & 1.000
        & 0.497 & 90.47 & 97.89 & 1.984 & 0.900 \\
        \rowcolor{gray!20}
        \textbf{Ours}
        & \textbf{0.646} & \textbf{89.49} & \textbf{97.55} & \textbf{1.910} & 0.645
        & \textbf{0.132} & \textbf{99.32} & \textbf{99.97} & \textbf{1.837} & \textbf{0.590} \\
        \midrule
        \multicolumn{11}{@{}l}{\textbf{3D tracking (EPE in metres)}} \\
        \midrule
        Track4World~\cite{lu2026track4world} + LI
        & 0.318 & 13.57 & 36.95 & 0.402 & 0.287
        & 0.284 & 19.26 & 42.86 & 0.365 & 0.255 \\
        \rowcolor{gray!20}
        TrackEverything~\cite{jain2026trackeverything} + LI
        & 0.417 & 18.01 & 49.39 & 0.586 & 0.357
        & 0.382 & 23.95 & 56.17 & 0.540 & 0.326 \\
        \textbf{Ours}
        & \textbf{0.166} & \textbf{38.83} & \textbf{71.29} & \textbf{0.342} & \textbf{0.104}
        & \textbf{0.148} & \textbf{47.42} & \textbf{73.92} & \textbf{0.305} & \textbf{0.093} \\
        \bottomrule
    \end{tabular}%
    }
\end{table*}

%% file: tables/tracking2d.tex
\begin{table}[t]
\centering
\scriptsize
\setlength{\tabcolsep}{3pt}
\renewcommand{\arraystretch}{1.15}
\caption{2D point tracking on HyperEvent and DSEC.
EPE2D is in pixels; AccS/AccR are percentages of errors strictly below
1/3 pixels. $\infty$ indicates failed predictions retained as infinite error.}
\label{tab:tracking2d}
\resizebox{\linewidth}{!}{%
\begin{tabular}{@{}l ccc ccc@{}}
\toprule
\textbf{Method}
& \multicolumn{3}{c}{\textbf{HyperEvent}}
& \multicolumn{3}{c}{\textbf{DSEC}} \\
\cmidrule(lr){2-4}\cmidrule(l){5-7}
& EPE2D $\downarrow$ & AccS $\uparrow$ & AccR $\uparrow$
& EPE2D $\downarrow$ & AccS $\uparrow$ & AccR $\uparrow$ \\
\midrule
EKLT~\cite{gehrig2019eklt}
& $\infty$ & 2.63 & 12.96
& $\infty$ & 8.98 & 16.02 \\
\rowcolor{gray!20}
DeepEvT~\cite{messikommer2023deepevt}
& 7.338 & 9.51 & 38.66
& 11.074 & 12.89 & 40.62 \\
TAPFormer~\cite{liu2026tapformer}
& 1.224 & 80.36 & 94.13
& 1.932 & 45.31 & 86.72 \\

\rowcolor{gray!20}
\textbf{Ours}
& \textbf{0.728} & \textbf{84.41} & \textbf{96.56}
& \textbf{1.605} & \textbf{55.86} & \textbf{89.84} \\
\bottomrule
\end{tabular}%
}
\end{table}

%% file: tables/full-3D.tex
\begin{table}[t]
\centering
\scriptsize
\setlength{\tabcolsep}{6pt}
\renewcommand{\arraystretch}{1.15}
\caption{EPE3D $\downarrow$ in metres for PointOdyssey/ADT and KITTI.
L-16/L-50 denote independent frame-0-to-target pair evaluations.
All the baselines use predicted geometry and all 50 clips per dataset while KITTI evaluations use 200 pairs.
$\infty$ denotes alignment failures retained in the evaluation.}
\label{tab:cross_dataset_epe3d}

\resizebox{\linewidth}{!}{%
\begin{tabular}{@{}l cc cc c@{}}
\toprule
\textbf{Method}
& \multicolumn{2}{c}{\textbf{PointOdyssey}}
& \multicolumn{2}{c}{\textbf{ADT}}
& \textbf{KITTI} \\
\cmidrule(lr){2-3}
\cmidrule(lr){4-5}
\cmidrule(l){6-6}
& \textbf{L-16} & \textbf{L-50}
& \textbf{L-16} & \textbf{L-50}
& \textbf{Pairs (200)} \\
\midrule
TrackEverything (Pi3)~\cite{jain2026trackeverything,wang2026pi3}
& 0.041 & 0.078
& $\infty$ & $\infty$
& 0.091 \\

\rowcolor{gray!20}
POMATO~\cite{zhang2025pomato}
& 0.266 & 0.319
& 0.075 & 0.124
& 0.234 \\

MASt3R~\cite{leroy2024mast3r}
& 0.339 & 0.514
& 0.206 & 0.200
& 0.519 \\

\rowcolor{gray!20}
MonST3R + RAFT~\cite{zhang2025monst3r,teed2020raft}
& 0.334 & 0.525
& 0.061 & 0.123
& 0.354 \\

ZeroMSF~\cite{liang2025zeromsf}
& 0.082 & 0.196
& 0.072 & 0.164
& 0.111 \\

\rowcolor{gray!20}
Track4World (DA3)~\cite{lu2026track4world,lin2025da3}
& 0.062 & 0.121
& 0.043 & 0.064
& 0.077 \\


\textbf{Ours}
& \textbf{0.047} & \textbf{0.091}
& \textbf{0.025} & \textbf{0.044}
& \textbf{0.076} \\
\bottomrule
\end{tabular}%
}
\end{table}

%% file: tables/imo.tex
\begin{table}[t]
\centering
\scriptsize
\setlength{\tabcolsep}{3pt}
\renewcommand{\arraystretch}{1.0}
\caption{IMO segmentation on HyperEvent (\%). mIoU averages
foreground/background IoU; pIoU is foreground IoU on pixel with at least one event.}
\label{tab:hyperevent_imo_metrics}

\begin{tabular}{@{}l cc@{}}
\toprule
\textbf{Method} &
\textbf{mIoU} $\uparrow$ &
\textbf{pIoU} $\uparrow$ \\
\midrule

\rowcolor{gray!20}
EVIMO \cite{wang2024unevimo} & 42.46 & 6.21 \\

Un-EVIMO \cite{wang2024unevimo} & 45.75 & 14.79 \\

\rowcolor{gray!20}
Event Suppression \cite{pellerito2026suppression} & 48.83 & 23.56 \\

\textbf{Ours} & \textbf{81.36} & \textbf{72.82} \\

\bottomrule
\end{tabular}
\end{table}

%% file: tables/ablations.tex
\begin{table*}[!t]
\centering
\scriptsize
\setlength{\tabcolsep}{2pt}
\renewcommand{\arraystretch}{1.0}
\caption{\textbf{Ablations at 5\,Hz RGB.}
A: Without IMO (T4W--TAPFormer); B: Ours;
C: ETAP Encoder (T4W--ETAP); D: Without Events (Ours--ZeroEvents).
Panels (a,c) evaluate RGB endpoints; (b) evaluates intermediate phases.
Panel (a) shows static, medium-motion, and high-motion points.
2D/3D denote EPE2D (px) / aligned residual EPE3D (m), both lower is better.
LI interpolates each model's own RGB endpoints; ours are event-conditioned.
IoU/F1 are fractions. Timing covers two-slot RGB inference with cached geometry,
not the full temporal path. Protocol details are in the supplementary.}
\label{tab:ablations}
\label{tab:ablation_motion}
\label{tab:motion_levels}
\label{tab:ablation_phase}
\label{tab:ablation_components}
\begin{minipage}[t]{0.32\textwidth}
\centering
\textbf{(a) GT motion levels}\par\smallskip
\begin{tabular}{@{}l cc cc cc@{}}
\toprule
\textbf{Method}
& \multicolumn{2}{c}{\textbf{Static}}
& \multicolumn{2}{c}{\textbf{Medium}}
& \multicolumn{2}{c}{\textbf{High}} \\
\cmidrule(lr){2-3}\cmidrule(lr){4-5}\cmidrule(lr){6-7}
& 2D $\downarrow$ & 3D $\downarrow$
& 2D $\downarrow$ & 3D $\downarrow$
& 2D $\downarrow$ & 3D $\downarrow$ \\
\midrule
A
& \textbf{0.341} & 2.968
& 1.968 & 2.772
& 16.576 & 3.269 \\
\rowcolor{gray!20}
\textbf{B}
& 0.525 & 0.077
& 1.653 & \textbf{0.138}
& \textbf{7.629} & \textbf{0.461} \\
C
& 0.560 & 7.556
& 2.355 & 4.644
& 10.590 & 4.370 \\
\rowcolor{gray!20}
D
& 0.378 & \textbf{0.057}
& \textbf{1.469} & 0.139
& 15.435 & 0.506 \\
\bottomrule
\end{tabular}
\end{minipage}\hfill
\begin{minipage}[t]{0.32\textwidth}
\centering
\textbf{(b) Inter-frame phase}\par\smallskip
\begin{tabular}{@{}l cc cc cc@{}}
\toprule
\textbf{Method}
& \multicolumn{2}{c}{$\phi=0.25$}
& \multicolumn{2}{c}{$\phi=0.50$}
& \multicolumn{2}{c}{$\phi=0.75$} \\
\cmidrule(lr){2-3}\cmidrule(lr){4-5}\cmidrule(lr){6-7}
& 2D $\downarrow$ & 3D $\downarrow$
& 2D $\downarrow$ & 3D $\downarrow$
& 2D $\downarrow$ & 3D $\downarrow$ \\
\midrule
A
& 1.508 & 0.814
& 2.463 & 2.275
& 2.925 & 2.882 \\
\rowcolor{gray!20}
\textbf{B}
& 1.366 & 0.082
& 2.072 & 0.120
& 2.521 & 0.147 \\
B + LI
& 1.174 & 0.048
& 1.744 & 0.084
& 1.757 & 0.111 \\
\rowcolor{gray!20}
C
& N/A & N/A & N/A & N/A & N/A & N/A \\
D
& 1.495 & 0.079
& 2.396 & 0.115
& 2.953 & 0.131 \\
\rowcolor{gray!20}
D + LI
& 1.040 & 0.047
& 1.731 & 0.081
& 2.195 & 0.109 \\
\bottomrule
\end{tabular}
\end{minipage}\hfill
\begin{minipage}[t]{0.32\textwidth}
\centering
\textbf{(c) Components and event input}\par\smallskip
\begin{tabular}{@{}l cccc@{}}
\toprule
\textbf{Metric} & A & \textbf{B} & C & D \\
\midrule
Overall 2D $\downarrow$ & 3.365 & 1.961 & 2.608 & 3.136 \\
\rowcolor{gray!20}
Overall 3D $\downarrow$ & 2.815 & 0.150 & 6.078 & 0.149 \\
IMO 2D $\downarrow$ & 6.550 & 3.472 & 4.765 & 6.040 \\
\rowcolor{gray!20}
IMO 3D $\downarrow$ & 2.654 & 0.227 & 4.521 & 0.245 \\
Static 2D $\downarrow$ & 0.341 & 0.525 & 0.560 & 0.378 \\
\rowcolor{gray!20}
Static 3D $\downarrow$ & 2.968 & 0.077 & 7.556 & 0.057 \\
IoU $\uparrow$ & N/A & 0.9565 & N/A & 0.9517 \\
\rowcolor{gray!20}
F1 $\uparrow$ & N/A & 0.9778 & N/A & 0.9752 \\
Params (M) & 1718.54 & 1719.34 & 1718.72 & 1719.34 \\
\rowcolor{gray!20}
Time (s) & 0.137 & 0.153 & 0.116 & 0.145 \\
\bottomrule
\end{tabular}
\end{minipage}
\end{table*}

%% file: sec/5_conclusion.tex
\section{Conclusion and Limitations}
\label{sec:conclusion}

We presented TrackAnyMotion, a monocular RGB--event tracker that recovers 2D and 3D point trajectories between RGB frames.
Its independent-motion feedback connects event measurements with scene geometry to guide iterative trajectory refinement.
We also introduced HyperEvent, providing synchronized frames, events, high-frequency trajectories, geometry, and independent-motion labels.
Together, these contributions advance point tracking beyond RGB sampling rates.
Reliance on monocular geometry and limited validation with real events remain limitations.

%% file: sec/X_suppl.tex
\clearpage
\maketitlesupplementary
\appendix
\renewcommand{\theequation}{S\arabic{equation}}
\renewcommand{\theHequation}{supp.\arabic{equation}}
\setcounter{equation}{0}
\newcommand{\ind}{\mathbf{1}}
\newcommand{\eps}{\varepsilon}

\section{Metric computation}
\label{sec:supp_metrics}

\paragraph{Notation and averaging.}
For source--target pair $(s,t)$ and material point $i$, the model
predicts source geometry $\widehat{\mathbf P}_{s,i}$, target image
position $\widehat{\mathbf u}_{t,i}$, and residual
$\widehat{\mathbf r}_{s\to t,i}$, with
$\widehat{\mathbf P}_{t,i}=\widehat{\mathbf P}_{s,i}
+\widehat{\mathbf r}_{s\to t,i}$.
Source and target XYZ coordinates use their respective camera frames;
$\mathbf r^{\rm gt}_{s\to t,i}=\mathbf P^{\rm gt}_{t,i}
-\mathbf P^{\rm gt}_{s,i}$ therefore includes camera motion.
Let $\Omega_{s,t}$ contain eligible correspondences and
$N=\sum_{s,t}|\Omega_{s,t}|$. We write
$\langle f\rangle=N^{-1}\sum_{s,t}\sum_{i\in\Omega_{s,t}}f_{s,t,i}$;
$\langle h\rangle_E$ averages over their $2N$ endpoints.
Tracking scores pool counts across pairs and scenes, including repeated
occurrences in overlapping windows. We use $\eps=10^{-6}$.

\paragraph{Geometry scale and alignment.}
Internally normalized geometry and residuals are restored using the
returned geometry scale before scoring. This restores backbone scale,
not absolute metric scale. For each method and pair, a shared positive
scale $a$ and translation $\mathbf b$ align both endpoint clouds:
\begin{equation}
 L(a,\mathbf b)=\sum_j
 \frac{\|a\widehat{\mathbf P}_j+\mathbf b-\mathbf G_j\|_1}
 {\|\mathbf G_j\|_2+\eps},\qquad a>0.
 \label{eq:supp_alignment}
\end{equation}
Here $\mathbf G_j$ is ground-truth XYZ. The released anchor-based
solver sets $\mathbf b=\mathbf G_k-a\widehat{\mathbf P}_k$, solves
for $a$ per anchor, and retains the lowest-loss valid fit. Dense
alignment uses mask-aware $32\times32$ endpoint samples; sparse
alignment uses all selected queries. No rotation is fitted.

\paragraph{2D tracking.}
With $e^{2D}=\|\widehat{\mathbf u}_t-\mathbf u_t^{\rm gt}\|_2$,
we report $\mathrm{EPE}_{2D}=\langle e^{2D}\rangle$ and
\begin{equation}
 \mathrm{Acc}_{2D}(\tau)=100\langle\ind[e^{2D}<\tau]\rangle,
\end{equation}
for $\tau\in\{1,3\}$ pixels. Material-grid coordinates are converted
to input-image pixels.

\paragraph{3D tracking.}
The shared translation cancels, giving
$e^{3D}=\|a\widehat{\mathbf r}_{s\to t}-\mathbf r^{\rm gt}_{s\to t}\|_2$
and $\mathrm{EPE}_{3D}=\langle e^{3D}\rangle$. With
$q=e^{3D}/(\|\mathbf r^{\rm gt}_{s\to t}\|_2+\eps)$,
\begin{equation}
 \mathrm{Acc}_{3D}(\tau)
 =100\langle\ind[e^{3D}<\tau\lor q<\tau]\rangle.
\end{equation}
AccS and AccR use $\tau=0.05$ and $0.10$, respectively, with
strict absolute-or-relative tests. Absolute errors are in meters when
GT is metric; KITTI uses its released, uncalibrated coordinate scale
for errors and thresholds. EPE is not normalized by depth, motion,
or elapsed time.

\paragraph{XYZ geometry.}
Define $\widetilde{\mathbf P}_j=a\widehat{\mathbf P}_j+\mathbf b$
and $d_j=\|\widetilde{\mathbf P}_j-\mathbf G_j\|_2$. Writing
$m_j=\min(\|\mathbf G_j\|_2,\|\widetilde{\mathbf P}_j\|_2)$,
we evaluate both endpoints as
\begin{align}
 \mathrm{AbsRel}_{XYZ}
 &=\left\langle\frac{d_j}{\|\mathbf G_j\|_2+\eps}\right\rangle_E,\\
 \delta_{1.25}^{XYZ}
 &=100\left\langle\ind[d_j<0.25m_j]\right\rangle_E.
\end{align}

\paragraph{Eligibility and regional scores.}
Native RGB-time scoring requires valid, visible, in-frame source and
target observations, finite GT XYZ, positive depth, and $s\ne t$.
Intermediate-time scoring requires valid material correspondences,
strict RGB bracketing, and non-query/non-padding timestamps; valid
occluded or out-of-frame targets can contribute. External datasets
use their specified GT masks. Predicted visibility and confidence do
not select the cohort. Native IMO/static subsets use valid source-frame
motion labels, exclude unknown labels, and reuse the overall pair's
alignment.

\paragraph{Independent-motion segmentation.}
Let $P$ and $G$ be predicted (threshold 0.5) and GT foreground sets,
$M$ the valid mask, and $E$ the nonzero signed-voxel support. Define
$J_M(A,B)=|A\cap B\cap M|/|(A\cup B)\cap M|$. Per-frame scores are
\begin{equation}
 \begin{aligned}
 \mathrm{mIoU}&=50\,[J_M(P,G)+J_M(P^c,G^c)],\\
 \mathrm{pIoU}&=100\,J_{M\cap E}(P,G).
 \end{aligned}
\end{equation}

\section{High-frequency trajectory interpolation}
\label{sec:supp_interpolation}

\paragraph{Protocol and baselines.}
RGB observations constrain track endpoints but do not specify the motion
between them. We evaluate trajectories at 200~Hz from RGB observations at
25~Hz and intervening events, predicting seven intermediate positions
per 40~ms interval without intermediate images. Native geometric
trajectories provide the evaluation targets. We compare our predictions
with linear interpolation of RGB-only tracks from Track4World and
TrackEverything, and with TAPFormer for 2D tracking.
The high-frequency tracking comparison in the main paper separates RGB observation timestamps
(S axis) from strictly bracketed intermediate timestamps (K axis), with
560,615 and 153,519 correspondence occurrences in the 2D evaluation, respectively.
Separate blocks report 2D endpoint error in pixels and 3D endpoint error
in metres, strict and relaxed accuracies, and regional EPEs for
independently moving objects and static regions. The 2D accuracy
thresholds are strictly below 1 and 3~pixels; 3D AccS/AccR use
absolute-or-relative thresholds of 0.05 and 0.10.
For an intermediate timestamp $t_a<t<t_b$, each linear 2D baseline predicts
$(1-\lambda)\widehat{\mathbf u}(t_a)+\lambda\widehat{\mathbf u}(t_b)$,
where $\lambda=(t-t_a)/(t_b-t_a)$ and both endpoints are predicted from
RGB observations.

\paragraph{2D tracking results.}
At intermediate timestamps (K axis), our method obtains an EPE of
0.132~px, compared with 1.470~px for Track4World + LI and
0.497~px for TAPFormer, corresponding to reductions of 91.0\% and
73.4\%, respectively. The fraction of predictions below 1~px increases
to 99.32\%, from 62.38\% and 90.47\%, while 99.97\% are below 3~px.
At RGB timestamps, our EPE is 0.646~px, a reduction of
47.0\% relative to the RGB baseline and 47.5\% relative to TAPFormer.
Thus, the improvement extends to observed timestamps, with a
larger relative EPE reduction at intermediate timestamps. This pattern
is consistent with the value of temporal information beyond the two
RGB endpoints. However, comparisons between different trained models
do not isolate the contribution of events or motion grounding; that
requires a matched no-event or component ablation.

\paragraph{Regional tracking results.}
Regional scores show lower intermediate-time errors in both regions:
our IMO EPE is 1.837~px, compared with 2.492~px for linear
interpolation and 1.984~px for TAPFormer, while static-region EPE decreases
to 0.590~px from 0.913 and 0.900~px, respectively. Static-region tracking
remains nontrivial because camera motion changes the image locations of
stationary world points. The gains therefore extend beyond independently
moving objects to points whose image motion is induced by the camera.

\paragraph{3D tracking results.}
For 3D interpolation, our K-axis EPE is 0.148~m, compared with
0.284~m for Track4World + LI and 0.382~m for TrackEverything + LI.
Our strict and relaxed accuracies reach 47.42\% and 73.92\%, respectively,
and both IMO and static-region EPEs are lower than those of either baseline.
The advantage also holds at RGB timestamps, where our EPE is 0.166~m
versus 0.318~m and 0.417~m, respectively.
TrackEverything's 2D EPEs are 2.695~px on S and 2.466~px on K.

\section{Ablation evaluation protocol}
\label{sec:supp_ablation_protocol}

The ablations compare 4 different models: T4W--TAPFormer (A), Ours (B),
T4W--ETAP (C), and Ours--ZeroEvents (D).

\paragraph{Sampling and temporal support.}
We retain every tenth native RGB frame, reducing the frequency from 50\,Hz to 5\,Hz. Native asynchronous events and the
200\,Hz trajectory grid remain available without regeneration.
The same three held-out synthetic scenes and source material points
are reused with seed 42. Each scene contains 256 fixed source points:
128 IMO and 128 static points. Input resolution remains
$256\times256$, and geometry settings are unchanged.

Each test window contains two retained RGB frames, separated by
200\,ms, and 41 native trajectory timestamps.
Eight RGB frames at 5\,Hz would require 1.4\,s, exceeding the native
clip duration; therefore, we use the smallest supported window.
Relative to the earlier eight-frame pilot, both RGB context length
and temporal coverage change. Comparisons between the two pilots
are consequently not frequency-only comparisons.

\paragraph{Alignment and aggregation.}
We use the EPE2D and aligned camera-residual EPE3D definitions in
Section~\ref{sec:supp_metrics}, with pixels and metres as their respective
units. The scale is fitted jointly to source and target points for each
method and pair using the scale-and-XYZ-shift alignment in
Eq.~\ref{eq:supp_alignment}; the shift cancels in the residual.
The same overall fit is reused for every motion group and IMO/static
region. Errors are pooled by correspondence counts without filtering
by predicted confidence.

\paragraph{Checkpoint settings.}
No additional training is performed. A/B/C reuse historical checkpoints
with 62k/88k/66k optimizer updates and effective batch sizes of 8/12/4.
They share RGB initialization and identical frozen geometry weights,
but differ in trainable components and supervision history.
A/B/D produce 41 joint temporal predictions, whereas C directly
predicts the two retained RGB slots.

\paragraph{Motion groups.}
Static points are identified using valid source IMO labels with
$\mathrm{mask}=0$. For moving points, we compute the norm of GT
world displacement divided by the elapsed time across the retained
RGB interval. This measures interval net-displacement speed.
Tertile boundaries, fixed from the selected GT cohort, are
$0.773189$ and $2.112327$\,m/s.
The RGB-endpoint cohort contains 423 correspondences: 217 static and
206 IMO, with 68, 69, and 69 in the low-, medium-, and high-motion groups.
The same cohort is used for the component comparison.

\paragraph{Temporal phase and eligibility.}
For a native timestamp $t$ between retained RGB timestamps $t_a$
and $t_b$, we define
\begin{equation}
\phi = \frac{t-t_a}{t_b-t_a}.
\end{equation}
The 200\,ms RGB interval contains 39 native intermediate timestamps.
We score all intermediate phases and report
$\phi\in\{0.25,0.50,0.75\}$ at the three phases shown in the main paper, corresponding to 50, 100, and 150\,ms
after the source frame.
Each reported phase contains 768 correspondences.

RGB-endpoint scoring requires known-visible, in-frame observations
with valid positive-depth geometry.
Intermediate scoring retains the existing GT-valid-state cohort,
including valid occluded and out-of-frame targets.

We additionally evaluate linear interpolation (LI) of each model's
own bracketing RGB predictions.
For our method, these endpoints are event-conditioned.
We interpolate image coordinates and camera-residual XYZ components
together with the model source pointmap; no world-space trajectory
interpolation is introduced.

\paragraph{Runtime and parameter counts.}
Inference time is measured on one GH200 GPU using the same
first-scene batch, two-slot RGB answer axis, four refinement
iterations, FP16, and cached frozen geometry.
We average three synchronized forwards after an explicit warmup,
excluding loading and preprocessing.
Parameter counts include frozen weights.

\section{Datasets}

\subsection{HyperEvent}

\subsubsection*{Synthetic Data Generation Pipeline}

We generate synthetic training data using a four-stage pipeline built on the
Kubric simulator with MOVi-F-style scenes: (1) physics simulation with
similarity time compression, (2) synchronized dual-rate rendering,
(3) event simulation, and (4) analytical generation of sparse and dense
tracking ground truth.

\paragraph{(1) Scene simulation and similarity time compression.}
Each scene is simulated using PyBullet rigid-body physics. Objects first
settle under normal gravity. We then apply \emph{similarity time compression}
with a factor of $k=4$: linear and angular velocities are scaled by $k$,
gravity by $k^2$, and the simulation and rendering rates by $k$. The
resulting trajectories satisfy $x'(t)=x(kt)$, meaning that the scene follows
the same trajectory but evolves $k$ times faster. This produces the fast
motion that motivates the use of event cameras, while preserving per-frame
pixel displacement, frame count, and rendering cost.

\paragraph{(2) Dual-rate rendering.}
We render two synchronized branches from the same physics simulation,
ensuring temporal correspondence between the event and frame data:

\begin{itemize}
    \item \textbf{High-rate (HR) branch.} Frames are rendered at
    $250k=1000$\,Hz and $256 \times 256$ resolution using a GPU
    rasterizer, with motion blur disabled. These frames are used as
    input to the event simulator.

    \item \textbf{Low-rate (LR) branch.} We render 50 RGB frames at
    30\,fps and apply motion blur in post-processing. We also export
    aligned depth maps as uint16 PNG images, with depth in meters
    decoded as $d = v/65535 \times 100$. These outputs are used as
    inputs to RGB baselines and for evaluation.
\end{itemize}

\paragraph{(3) Event simulation.}
We generate events from the HR frames using the CUDA backend of the Event
Simulator (ESIM). Each frame is converted to log-luminance,
$L=\log(0.299R+0.587G+0.114B)$. Between consecutive HR frames, log-luminance
is assumed to vary linearly. An event is emitted when the interpolated
log-luminance crosses a contrast threshold of $\pm 0.2$ relative to the
per-pixel reference. Event timestamps are obtained by linear interpolation
and stored in microseconds. The resulting events are sorted by timestamp.

\paragraph{(4) Ground-truth tracks and correspondences.}
We generate all labels analytically from the simulator state, including
rigid-body poses and mesh geometry:

\begin{itemize}
    \item \textbf{Sparse tracks.} We sample 384 points per scene from
    mesh surfaces and record their world-space 3D positions, pixel
    projections, visibility, instance identities, and dynamic or
    static labels. Under the Blender camera convention, depth is
    defined as $-z_{\mathrm{cam}}$.

    \item \textbf{Dense geometry.} We export dense geometry at 250\,Hz,
    which is an integer divisor of the 1000\,Hz HR rendering rate. Thus,
    every dense timestamp coincides with a rendered frame and requires
    no temporal interpolation. Each dense frame contains depth and a
    per-pixel surface identity buffer comprising the instance ID,
    triangle ID, and perspective-correct barycentric weights. Both are
    stored at full resolution ($256 \times 256$) and at $1/8$ spatial
    resolution. Nearest-neighbor subsampling preserves discrete IDs.

    \item \textbf{Dense correspondences.} We derive 2D and 3D scene
    flow, target coordinates, and four-state labels
    (visible, occluded, out-of-frame, or invalid) analytically on
    demand from the identity buffers. Each surface point is reconstructed
    in object space, transformed into world coordinates at the source
    and target times using the stored object poses, and reprojected into
    the target view. Occlusion is determined by a depth test against
    the target depth reconstructed from the target pixel's own identity
    buffer.
\end{itemize}

\paragraph{Data splits.}
We generate datasets in tiers using deterministic seeds and partition
them into training, validation, and test sets with an 80/10/10 split.

\subsection{Real-World Dataset}

To construct a real-world dataset, we capture several fast-moving objects
using a calibrated and synchronized setup comprising a Prophesee Gen4 event
camera and a FLIR Blackfly S color camera operating at 10\,fps. The dataset
has a spatial resolution of $945 \times 649$ and includes accurate camera
poses provided by a motion capture system. We use this dataset to
qualitatively evaluate our method on real-world scenes.



\subsection{Time Lens Dataset}
\label{sec:supp_timelens}

The High-Speed Event and RGB (HS-ERGB) dataset, introduced with
Time Lens~\cite{tulyakov2021time}, contains real event streams and
high-speed color videos of dynamic indoor and outdoor scenes.
The recording setup combines a Prophesee Gen4 event camera
($1280\times720$) with a global-shutter FLIR Blackfly S RGB camera
($1440\times1080$), separated by a 2.5\,cm baseline and synchronized
in hardware. The recorded sequences use RGB frame rates of
150--163\,Hz. They comprise close-range scenes with fast, nonlinear
object motion and distant scenes dominated by camera motion,
including spinning objects, bursting balloons, water splashes,
and rapid camera rotations.

Spatial alignment uses stereo rectification, with an additional
feature-based homography for close-range planar scenes recorded
with a stationary camera. HS-ERGB complements our synthetic benchmark
with real sensor recordings of challenging motion. Its high-rate RGB
frames provide image references for frame-interpolation evaluation;
the original release does not include annotated 2D/3D point
trajectories or independent-motion masks. It therefore supports
qualitative assessment of our tracking and segmentation without
providing ground truth for their endpoint errors or mask IoU.